\documentclass[letterpaper, 10 pt, conference]{ieeeconf}  

\IEEEoverridecommandlockouts                              

\usepackage{balance}

\usepackage{amsmath} 
\usepackage{amssymb}  

\usepackage[normalem]{ulem}
\usepackage{multirow, booktabs, makecell, float}
\usepackage{gensymb}
\usepackage[table,xcdraw]{xcolor}

\usepackage{pifont,xcolor}
\usepackage{graphicx}
\usepackage{bm}
\usepackage{xcolor}
\usepackage{cite}
\usepackage{makecell}

\usepackage[font=footnotesize]{caption}
\usepackage[hidelinks]{hyperref}

\title{\LARGE \bf
MM-TRELLIS: Point-Cloud Guided Multi-Modal 3D Vehicle Generation in Autonomous Driving 
}

\author{
    Hongli Xiao$^{1,2,3*}$, Youjian Zhang$^{3*}$,
    Yucai Bai$^{3}$, Chaoyue Wang$^{5}$, Yaohui Jin$^{1}$, \\
    Xiaoguang Ren$^{2}$,  Wenjing Yang$^{4}$, Long Lan$^{4\dagger}$
\thanks{*Equal contribution. Work done during internship at Bosch.}
\thanks{$\dagger$Corresponding author: Long Lan (long.lan@nudt.edu.cn).}
\thanks{$^{1}$MoE Key Lab of Artificial Intelligence, AI Institute, Shanghai Jiao Tong University, China (honglixiao@sjtu.edu.cn)}%
\thanks{$^{2}$Academy of Military Science, China }%
\thanks{$^{3}$Bosch innovation software development (Wuxi) Co., Ltd., China Technology, China (Youjian.Zhang@cn.bosch.com)}%
\thanks{$^{4}$College of Computer Science and Technology, National University of Defense Technology, China}%
\thanks{
$^{5}$Shopee Pte. Ltd., China
}%
}

\begin{document}

\maketitle
\thispagestyle{empty}
\pagestyle{empty}

\begin{abstract}

Recovering realistic 3D vehicle models from autonomous driving scenes is crucial for synthesizing training data  and building simulation environment.
However, most existing vehicle generation methods fail to fully exploit multimodal sensors (\textit{i.e.} multi-view images and LiDAR point clouds) and rely on neural rendering based reconstruction, leading to low-quality mesh.
Recently, native 3D generative models have made significant progress, yet they are not built for arbitrary multi-view inputs and often struggle with in-the-wild driving images.
In this work, we present MM-TRELLIS, a multi-modal version of TRELLIS for in-the-wild 3D vehicle generation that integrates LiDAR and image sensors from autonomous driving datasets into native 3D generative models.
Specifically, multi-view images are cycled as conditioning inputs, while LiDAR point clouds provide test-time guidance to ensure geometric accuracy and cross-view consistency.
During denoising, we first align the guidance point cloud with the model priors, then enforce consistency between the generated geometry and the guidance point cloud.
Finally, we introduce a voxel filtering strategy based on  the opacity of 3D Gaussian Splatting  to suppress floaters and produce clean meshes.
Comprehensive experiments on Waymo dataset demonstrate our method outperforms existing methods in high-fidelity 3D vehicle generation. 
{ Code is available at \href{MM-TRELLIS}{https://github.com/HongliXiao/MM-TRELLIS}.}

\end{abstract}

\section{INTRODUCTION}

The development of autonomous driving and robotic systems increasingly relies on large-scale and diverse 3D assets. In particular, high-fidelity vehicle models are crucial for synthesizing training data for perception modules, constructing realistic simulation environments, and enabling safe interaction in complex urban scenes. 
Nevertheless, manual creation is costly and unscalable, calling for automatic approaches to recover 3D vehicle models directly from raw sensory data.
In this work, we intend to generate high-fidelity 3D vehicle assets from existing multimodal self-driving datasets~\cite{sun2020waymo}.

Recent efforts have explored creating 3D vehicle models in real-world road scenarios, yet fundamental limitations remain. 
Methods that reconstruct solely on single-view images~\cite{muller2022autorf, guo2024supnerf,liu2024carstudio, lin2024drive123} are inherently ill-posed and tend to produce ambiguous geometry. 
Multi-view methods~\cite{xu2023discoscene, du2024dreamcar} mitigate some of the ill-posedness, but the limited view-points and severe occlusions in driving datasets still lead to incomplete reconstructions and degraded fidelity. 
Moreover, most vehicle generation methods~\cite{shen2023gina3d, liu2024carstudio,du2024dreamcar, liu2025stnerf, xu2023discoscene, yang2025genassets } rely on neural rendering–based reconstruction, leading to slow inference and low-quality mesh extraction.

\begin{figure}[t]
      \centering
      
      \includegraphics[width=1.0\linewidth]{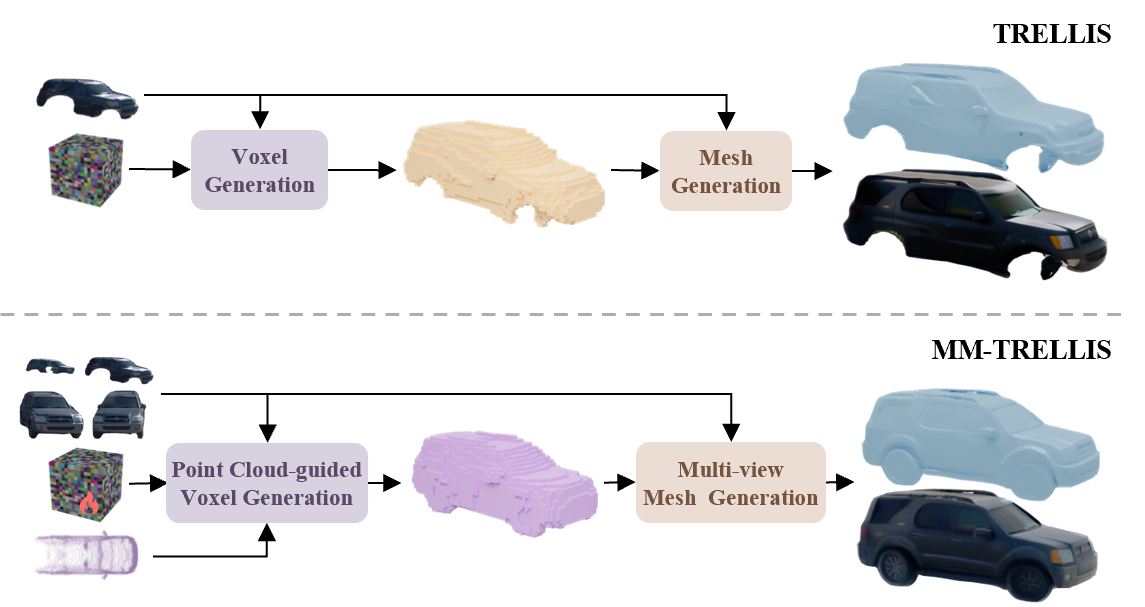}
      \caption{  TRELLIS vs.\ MM-TRELLIS.
      Top (TRELLIS): single-image voxel generation followed by mesh decoding.
    Bottom (MM-TRELLIS): We propose a point cloud–guided voxel generation and mesh generation with multi-view inputs, yielding robust and accurate geometry.
      }
      \label{fig:teaser}
      \vspace{-6mm}
\end{figure}


Recently, native 3D generative models that directly generate 3D representations~\cite{zhang2024clay, zhao2025hunyuan3d2.0, lai2025hunyuan3d2.5, xiang2025trellis} have achieved remarkable progress in synthesizing high-quality 3D assets.
Trained on large-scale synthetic datasets~\cite{chang2015shapenet, deitke2023objaverse, deitke2023objaversexl}, these models learn strong shape priors and show impressive generalization ability to unseen image textures.
Among them, TRELLIS~\cite{xiang2025trellis} represents a milestone by introducing a two-stage generation pipeline with structured voxel latents, where the first stage predicts a coarse geometric structure in a sparse 3D grid, and the second stage enriches it with local latent features. 
However, TRELLIS is not specifically designed for in-the-wild 3D vehicle generation, and several limitations arise when directly applying it to this task:
i) the generative model is trained with single-image conditioning and tends to produce inaccurate geometry when handling multi-view inputs. In addition, it lacks the ability to incorporate LiDAR point clouds;
ii) the framework assumes high-resolution, object-centered images as input, and its performance degrades significantly when conditioned on lower-quality inputs with diverse viewpoints and heavy occlusions (as shown in Fig.~\ref{fig:teaser}).


Our goal is to leverage the strong 3D geometric priors of state-of-the-art native 3D generation methods while fully exploiting the multimodal inputs specialized for autonomous driving datasets.  Thus, we present MM-TRELLIS, a zero-shot 3D vehicle generation method with multi-modal inputs. 
First, given multiple viewpoints of a vehicle, TRELLIS can cyclically inject different viewpoints as conditioning inputs to aggregate complementary visual information. However, naively adopting such cycle-conditioning is suboptimal, as different image conditions may provide contradictory optimization gradient, leading to inaccurate and low-quality geometry.
To address this issue, we incorporate LiDAR point clouds as test-time guidance to encourage consistency between the generated voxel and the LiDAR structure. By leveraging the iterative nature of the diffusion process, our point cloud guidance scheme provides explicit geometric supervision without retraining the model, thereby avoiding the risk of degrading the learned 3D geometric prior.

Noted that the generated voxel does not have a fixed orientation (varies with the conditioning view), so we first optimize the orientation of the guidance point cloud to align with the generated voxel in order to make the guidance work properly.
Finally, there are still floaters and extra geometry in the mesh generation, therefore, we propose a voxel filtering strategy based on the opacity of generated 3D Gaussian Splatting (3DGS)~\cite{kerbl20233dgs} to further refine and clean the geometry.

Our main contributions are summarized as follows:
\begin{itemize}
\item We introduce MM-TRELLIS, a zero-shot 3D vehicle generation framework that adapts native 3D diffusion priors to multimodal autonomous driving data. 
\item We propose a point cloud-guided test-time optimization scheme. 
By resolving the orientation disparity and regularizing the geometry generation with LiDAR structure during denoising, it imposes explicit geometric supervision and effectively enhances robustness under occlusions and challenging viewpoints.
\item We propose an opacity-based voxel filtering strategy to remove floaters and further improve surface fidelity.
\item Extensive experiments show that MM-TRELLIS surpasses existing 3D/vehicle generation methods, enabling high-quality in-the-wild 3D vehicle generation.

\end{itemize}

\section{RELATED WORK}

\subsection{3D Vehicle Modeling in the Wild}

Recovering 3D vehicle models from in-the-wild driving data has been widely studied~\cite{muller2022autorf, niemeyer2021giraffe, chan2022eg3d, karpikova2025madrive, li2025refsam, xia2025d, lan2026c}.  
Single-view methods~\cite{liu2024carstudio, shen2023gina3d, guo2024supnerf, lin2024drive123} take one image as input and often rely on category priors or large-scale training.  
While effective in constrained settings, they remain ill-posed in driving scenarios with occlusions and challenging viewpoints, leading to geometric ambiguity and inconsistent novel views.  
Multi-view approaches~\cite{xu2023discoscene, du2024dreamcar} mitigate ambiguities by aggregating multiple observations, but the limited camera viewpoints and frequent occlusions in driving datasets still result in incomplete reconstructions and degraded fidelity.  
Approaches that incorporate LiDAR take different forms: GINA-3D~\cite{shen2023gina3d} uses LiDAR depth as auxiliary supervision during training, GenAssets~\cite{yang2025genassets} learns latent priors with rendered LiDAR constraints, and ProtoCar~\cite{liu2025protocar} leverages completed LiDAR point clouds only as training labels. In all cases, LiDAR serves as supervision rather than directly guiding generation, which limits their ability to impose reliable geometric constraints.
In contrast, our method employs LiDAR as test-time guidance tightly coupled with the denoising trajectory, allowing it to directly guide geometry generation under occlusions and challenging viewpoints.

\subsection{Native 3D Generative Models}

Early work~\cite{chen2023fantasia3d, qianmagic123, lin2023magic3d, pooledreamfusion, wang2024prolificdreamer} in 3D generation often relied on optimization-based pipelines that lifted 2D priors into 3D representations, but these methods are slow and limited in fidelity. 
More recently, feed-forward paradigms have enabled direct 3D synthesis, comprising 2D-lifting methods~\cite{liu2023zero123, long2024wonder3d, shi2023zero123++, wu2024unique3d, xu2024instantmesh} that exploit image priors and native 3D models~\cite{zhang20233dshape2vecset, zhang2024clay, xiang2025trellis, zhao2025hunyuan3d2.0, hunyuan3d2025hunyuan3d2.1, lai2025hunyuan3d2.5} that learn directly in 3D space.

Among them, native 3D generative models have shown the strongest potential for high-fidelity and scalable synthesis. Trained on large datasets, these models learn powerful shape priors and support controllable generation across categories. 
Representative designs include latent set representations~\cite{zhang20233dshape2vecset, chen2025dora, zhao2023michelangelo, zhang2024clay}, voxel-structured latents~\cite{xiang2025trellis, ye2025hi3dgen}, and scalable sparse encodings~\cite{li2025sparc3d, wu2025direct3ds2}. In particular, TRELLIS~\cite{xiang2025trellis} introduced a two-stage pipeline where coarse voxel grids are first generated and then enriched with local features, significantly improving scalability, stability, and interpretability.
Despite these advances, existing models are primarily trained on clean synthetic assets and conditioned on single-view images or text prompts. This leaves a notable domain gap when applied to autonomous driving data, where vehicles appear under occlusions, extreme viewpoints, and sensor noise, limiting their applicability to real-world vehicle modeling.  


\section{Method}

\begin{figure*}[t]
      \centering
      \includegraphics[width=1.0\linewidth]{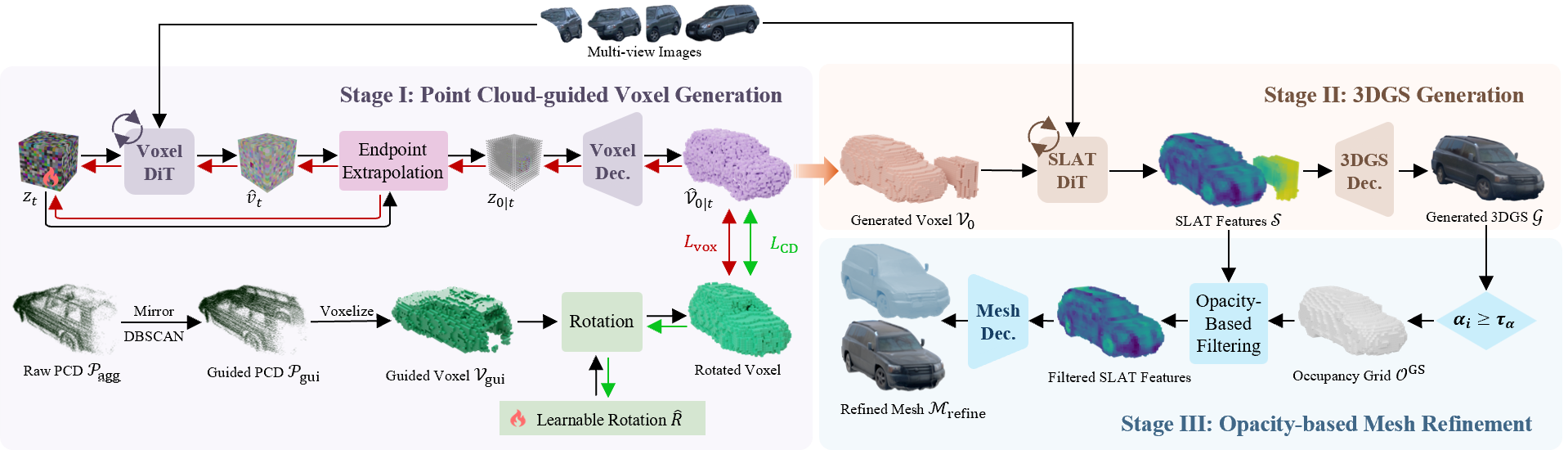}
      \caption{ 
      \textbf{Overview of proposed MM-TRELLIS.} 
      In stage I, we generate voxels with multi-view cycle conditioning and point cloud guidance. The LiDAR point cloud is first preprocessed and rotated by a learnable parameter $\hat{R}$ to a aligned orientation. Then the voxel guidance is applied to optimize the sampled latent during the denoising process.
      Stage II perform a 3DGS generation with multi-view conditioning, ensuring the texture fidelity of the generation.
      Finally, opacity-based mesh refinement is performed in Stage III: a voxel mask is obtained by thresholding Gaussian opacity and filtering SLAT features, and decoding the filtered features produces the final clean mesh.
      }
      \label{fig:pipe}
      \vspace{-2mm}
\end{figure*}

We aim to generate realistic 3D vehicle assets from multimodal data based on native 3D generation method. The overall framework is illustrated in Fig.~\ref{fig:pipe}.
Given multi-view images of a vehicle, MM-TRELLIS performs cycle-conditioning to aggregate complementary appearance cues across different viewpoints. Meanwhile, LiDAR point clouds are incorporated as a test-time guidance signal, ensuring an accurate and synchronized geometry. 
Finally, an opacity-based voxel filtering step is applied to suppress floaters and produce clean, high-fidelity 3D vehicle models.


\subsection{Multi-view Cycle Conditioning}

In autonomous driving datasets, vehicles are typically captured from different onboard cameras and across multiple frames, offering complementary visual observations for 3D generation. 
While a single image often suffers from occlusion, truncation, or perspective distortion, aggregating information across views can provide richer and more complete appearance information. 
Although not explicitly stated in the original TRELLIS paper~\cite{xiang2025trellis}, a straightforward way to exploit such multi-view information is to employ a cycle-conditioning mechanism.
Formally, given $N$ selected images of a vehicle:
\begin{equation}
\mathcal{I} = \{ I_1, I_2, \dots, I_{N} \},
\end{equation}
where each $I_i$ may come from different cameras or different frames.  During iterative denoising, the diffusion model predicts a cleaner latent sample $\mathbf{x}_t$ from noisy input $\mathbf{x}_{t+1}$ conditioned on an image $I_i \in \mathcal{I}$. 
We perform cycle-conditioning by switching the conditioning image at each step:
\begin{equation}
\mathbf{x}_t = f_\theta(\mathbf{x}_{t+1}, I_{(t \bmod N) + 1}, t),
\end{equation}
where $f_\theta$ is the denoising network, and the index cycles over the available views. This ensures that each denoising step is informed by a different viewpoint of the same vehicle.



\subsection{LiDAR point cloud-guided Geometry Generation}


Simply applying multi-view image conditioning can be suboptimal: i) images in autonomous driving scenes often suffer from occlusions and varying viewpoints, leading to incomplete or distorted geometry; ii) multi-view conditions may not cooperate effectively, as they can provide contradictory optimization gradients at each timestep.



To address these issues, we involve LiDAR point clouds as a test-time guidance signal for the denoising process of the voxel generation. 
Unlike RGB images, LiDAR observations can be easily aggregated across multiple frames, providing metric-accurate surface measurements that remain reliable under occlusions, varying illumination, and long-distance viewpoints. Consequently, LiDAR point clouds can serve as an effective modality for accurate and robust geometric constraints. Specifically, the point-cloud guidance can be decomposed into three stages: i) LiDAR preprocessing; ii) Voxel-guided geometry generation; iii) Voxel pose optimization, as detailed in the following.


\noindent \textbf{LiDAR Preprocessing.}  
Raw LiDAR point clouds collected in one timestep often contain noise, outliers, and missing regions due to sensor accuracy error and occlusions.
To produce an accurate and complete reference for guidance, we first aggregate the raw vehicle point cloud $\mathcal{P}_{\text{agg}} = \sum_{i=0}^{N} \mathcal{P}_i$ through the sequence, while $\mathcal{P}_i$ is the point of target vehicle in each frame. 

However, in driving scenes, a vehicle is usually observed from only one side, leaving the opposite side of the point cloud empty. Thus, we exploit the symmetry characteristic of vehicles by mirroring the raw point cloud across the vehicle’s center plane, producing a symmetrized point set:
\begin{equation}
\mathcal{P}_{\text{sym}} = \mathcal{P}_{\text{agg}} \cup  {Mirror}(\mathcal{P}_{\text{agg}}).
\end{equation}
The $Mirror(\cdot)$ operation flips the points by left-right symmetry plane, which can be estimated from bounding boxes.

We randomly downsample the point cloud if it contains more than $100{,}000$ points, in order to preserves sufficient geometric fidelity while ensuring computational efficiency:
\begin{equation}
\mathcal{P}_{\text{down}} \subset \mathcal{P}_{\text{sym}}, \quad
|\mathcal{P}_{\text{down}}| = \min(|\mathcal{P}_{\text{sym}}|, 10^5).
\end{equation}


Finally, we apply DBSCAN~\cite{ester1996dbscan} clustering to filter out scattered outliers, which can be formulated as:
\begin{equation}
\mathcal{P}_{\text{gui}} = \{ \mathbf{p} \in \mathcal{P}_{\text{down}} \mid \text{DBSCAN}(\mathcal{P}_{\text{down}})_\text{label}(\mathbf{p}) \neq -1 \},
\end{equation}
where points with label $-1$ are considered outliers in DBSCAN clustering results and are removed. This step effectively removes background noise and irrelevant points, resulting in a clean point cloud $\mathcal{P}_{\text{gui}}$ as final guidance.


\noindent \textbf{Voxel-guided Geometry Generation.}  
One possible way to exploit point clouds as a conditioning signal is to encode the point cloud into a latent embedding and train cross-attention layers to impose the condition. However, this approach requires tedious post-training of the DiT. Furthermore, real-world 3D vehicle datasets are extremely scarce compared to the large-scale datasets used for training native 3D generative models. Post-training on such small datasets risks degrading the strong and generalizable 3D geometric priors.

Inspired by Marigold-DC~\cite{viola2024marigold_dc}, which formulates depth completion as monocular depth prediction guided by a sparse depth map, we aim to guide TRELLIS’s voxel generation towards the shape of a reference point cloud at test time without re-training model parameters. 


First, we want to obtain both the guidance voxel and the predicted voxel in timestep $t$. Specifically, we voxelize $\mathcal{P}_{\text{gui}}$ into a voxel grid ${\mathcal{V}_{\text{gui}}} \in \{0,1\}^{H \times W \times D}$, where $\mathcal{V}_{\text{gui}}(u,v,w)=1$ if the voxel in coordinate $(u,v,w)$ contains at least one LiDAR point and $0$ otherwise.  
Meanwhile, at inference timestep $t$, the flow-matching DiT estimates a vector field $\hat{v}_t$, 
and we ``preview'' the denoised latent via an endpoint extrapolation~\cite{lipman2022flowmatching, choi2025enhanced}:
\begin{equation}
\begin{aligned}
z_{0\mid t} &= z_t - t \hat{v}_t, \\
\end{aligned}
\end{equation}
where $z_t$ is the noisy latent at timestep $t$, $z_{0\mid t}$ is the corresponding estimated denoised result.
Decoding $z_{0\mid t}$ with the voxel decoder yields the predicted occupancy $\hat{\mathcal{V}}_{0|t}$.



Then we calculate a masked Binary Cross Entropy (BCE) loss between the reference voxel grid ${\mathcal{V}_{\text{gui}}}$ and the predicted voxel output $\hat{\mathcal{V}}_{0|t}$ only on the occupied voxels in ${\mathcal{V}_{\text{gui}}}$:
\begin{equation}
\mathcal{L}_{\mathrm{vox}}(t) =
- \frac{1}{|\mathcal{V}_{\text{gui}}|}
\sum_{(u,v,w)} \mathcal{V}_{\text{gui}}(u,v,w)\,\log \hat{\mathcal{V}}_{0|t}(u,v,w),
\end{equation}
where $|\mathcal{V}_{\text{gui}}|$ counts the nonzero voxels, and $(u,v,w)$ is the 3D coordinates of the voxel grid.

By restricting supervision to non-empty voxels, we avoid penalizing regions without point cloud observations, enabling the use of incomplete point clouds as guidance. The gradients of the voxel guidance loss are back-propagated to the sampled latents, thereby steering the flow-matching process.


Notably, the final voxel generation is jointly conditioned on both the point cloud and input images. In other words, while the voxels are regularized to align with the reference point cloud, TRELLIS’s inherent 3D prior complements the missing regions. This leads to outputs with metric-accurate geometry while effectively addressing occlusions.


\noindent \textbf{Voxel Pose Optimization.}
In early experiments, we observed that directly applying the voxel guidance loss sometimes led to poor convergence. This is because the orientation of voxels generated under the image condition is ambiguous and does not necessarily align with the canonical orientation of the reference voxel. When these two optimization objectives become too contradictory, the convergence becomes difficult.

To stabilize training, we optimize the pose of the guided voxel $\mathcal{V}_{\text{gui}}$ such that it approximately aligns with the predicted coarse voxel. 
As the rectified-flow model holds a nice property that the sampling trajectory follows a stable probability transitions \cite{lipman2022flowmatching}, empirically, the intermediate voxel output at around timestep $t=0.9$ already exhibits geometry similar to the final result. 
This property allows us to first adjust the voxel orientation during the early denoising steps, after which voxel-guided geometry generation can proceed with well-aligned voxels.

Specifically, we convert both reference voxel and predicted voxel into point cloud:
\begin{equation}
\begin{aligned}
P &= \{\, p \in \mathbb{R}^3 \;\mid\; \mathcal{V}_{\text{gui}}(p) = 1 \,\}, \\
Q &= \{\, q \in \mathbb{R}^3 \;\mid\; \hat{\mathcal{V}}_{0|t}(q) = 1 \,\}.
\end{aligned}
\end{equation}
Then we define a learnable rotation parameters as a 3 dimension anxis angles, denoted as $R \in SO(3)$, and minimizing the one-sided Chamfer Distance (CD) between the predicted point cloud and the reference point cloud after rotation:
\begin{equation}
    \mathcal{L}_{\mathrm{CD}}(P, Q) 
= \frac{1}{|P|} \sum_{p \in P} \min_{q \in Q} \|Rot(p,R) - q\|_2^2,
\end{equation}
where $Rot(\cdot,\cdot)$ is a rotation transformation operator. We only calculate one-sided Chamfer Distance since the reference voxel can be incomplete. The gradient of $\mathcal{L}_{\mathrm{CD}}$ is back-propagated to the rotation parameters $R$, thus optimize the orientation for voxel guidance.


In summary, our full test-time guidance uses a three-phase schedule within a 100-step sampler:
i) Early sampling: run Stage I DiT and pause at step 30 ($t{\approx}0.88$) to decode a coarse voxel;
ii) Pose alignment: optimize a rotation $R$ (applied to the guidance voxel grid $\mathcal{V}_{\text{gui}}$) for 100 iterations by minimizing $\mathcal{L}_{\mathrm{CD}}$;
iii) Guided generation: resume denoising to $t{=}0$ by minimizing $\mathcal{L}_{\mathrm{vox}}$ at each step.

\subsection{Opacity-based Voxel Filtering}


Since the voxel guidance loss constrains only on voxels containing LiDAR points, it is not sufficient to suppress the generation of extra voxels, which may lead to floaters and noisy surface in the final generated mesh. 

Interestingly, despite the presence of floaters in the mesh generation, TRELLIS's  Gaussian Splatting (3DGS) generation still produces clean renderings that faithfully match the multi-view image conditions. We infer that this robustness arises from the strong texture prior provided by multi-view images, as SLAT features are originally trained from the back-projection of image features. Leveraging this property, we perform voxel filtering based on the opacity of 3DGS.

Let $\mathcal{{S}}$ be the generated SLAT features from TRELLIS stage II,
decoding them as 3DGS yields a set of Gaussians
\begin{equation}
\mathcal{G} \;=\; \bm{\mathcal{D}}_{\text{GS}}(\mathcal{S})
\;=\; \{ (\mu_i,c_i, s_i,\alpha_i,r_i) \}_{i=1}^{K},
\end{equation}
where $\mu_i,c_i, s_i,\alpha_i,r_i, o_i$ denotes for positions, colors, scales, opacities, and rotations respectively.
As aforementioned, the floaters are not visible in the rendering results, indicating the opacity of these regions are very low. Therefore, we first prune low-opacity splats through a threshold $\tau_\alpha$:
\begin{equation}
\mathcal{G}_{\text{prune}} \;=\; \{\, g_i \in \mathcal{G} \mid \alpha_i \ge \tau_\alpha \,\}.
\end{equation}
The retained Gaussians $\mathcal{G}_{\text{prune}}$ are voxelized into an occupancy grid $\mathcal{O}^{\text{GS}}\!\in\!\{0,1\}^{H\times W\times D}$, which is used as a mask for filtering the SLAT feature grid $\mathcal{S}$, 
and the refined mesh is generated using the TRELLIS mesh decoder:
\begin{equation}
\mathcal{M}_{\text{refine}} 
= \bm{\mathcal{D}}_{\text{mesh}}\!\left(\mathcal{S} \odot \mathcal{O}^{\text{GS}}\right).
\end{equation}

Similar to voxel-guided geometry generation, this procedure is a test-time refinement that does not require re-training the model parameters. We leverage the robustness of Gaussian Splatting generation to in turn refine the generated mesh, resulting in cleaner and high-fidelity mesh generations for downstream tasks.


\begin{table*}[t]
\centering
\caption{ 
Quantitative comparison on Waymo dataset. The quality of Novel-view Synthesis is measured by PSNR/SSIM/LPIPS.
\protect\\
The accuracy of geometry is measured by Chamfer Distance (CD$_{one}$: LiDAR$\rightarrow$mesh, CD$_{bi}$: LiDAR$\leftrightarrow$mesh). 
}
\setlength{\tabcolsep}{6pt}
\renewcommand{\arraystretch}{1.2}
\label{tab:comp1}
\begin{tabular}{c|l|cccccc}
\toprule
Image input & Method & PSNR $\uparrow$ & SSIM $\uparrow$ & LPIPS $\downarrow$ & CD$_{one}\downarrow$ & CD$_{bi}\downarrow$  & Time $\downarrow$ \\
\midrule
\multirow{2}{*}{Single-view  }
   & InstantMesh~\cite{xu2024instantmesh}   &	   22.34    &  0.9218	  &  0.0843  &   0.1189  &	 0.1892 & 9.6s \\
 & TRELLIS~\cite{xiang2025trellis}   &	19.81   &  0.9037	  &  0.0912 &    0.0848 &	0.1050 & 25.8s\\
\midrule
\multirow{4}{*}{Multi-view }
 &  DreamCar~\cite{du2024dreamcar}  	&    20.19	& 0.9022 &	0.0972	& 0.1491 &	0.2128  & 118min\\
 &  MV-Hunyuan3D~\cite{zhao2025hunyuan3d2.0} 	&   22.45		& \textbf{0.9231}		& 0.0839		& 0.2443	& 	0.1318  & 13.7min \\
 &  MV-TRELLIS~\cite{xiang2025trellis}   &   20.39		& 0.9081		& 0.0834		& 0.0388		& 0.0479 & 24.3s \\
 & \textbf{Ours}        	&    \textbf{22.61}	& 	0.9159		& \textbf{0.0720}		& \textbf{0.0210}		& \textbf{0.0361} & 37.3s\\
\bottomrule

\end{tabular}

\end{table*}

\begin{figure*}[t]
      \centering
      \includegraphics[width=1.0\linewidth]{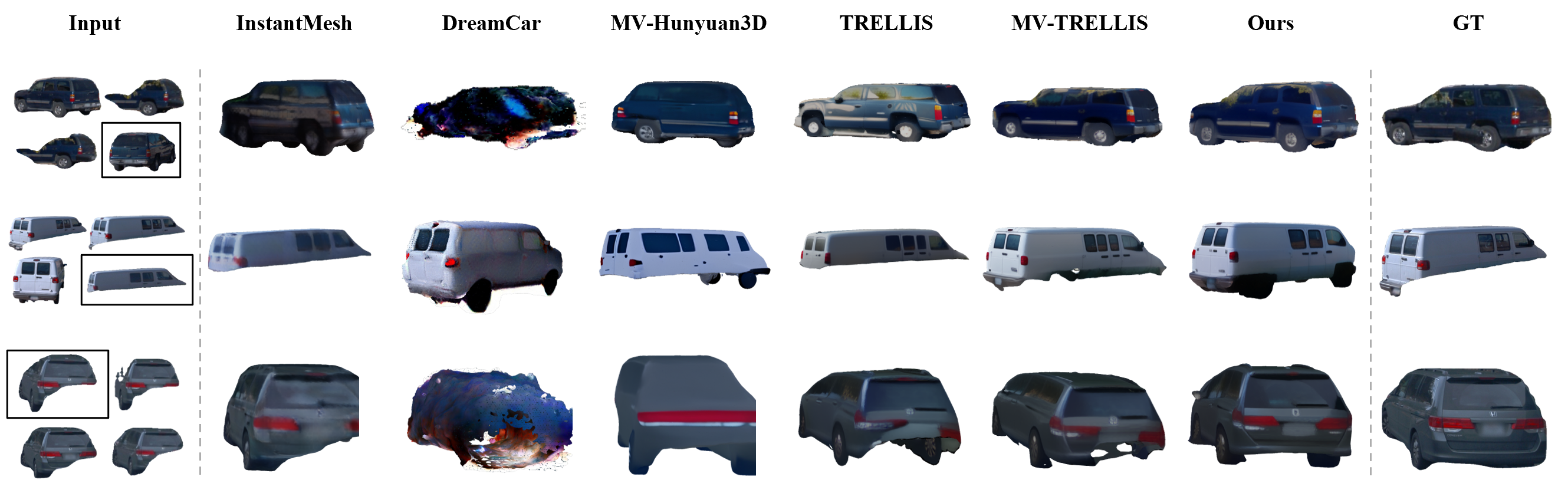}

      \caption{ Qualitative comparison with baseline methods in novel view synthesis.
      The four images in Input column represent the multi-view inputs for multi-view methods, while the image in the black box is used for the single-view method.
      }
      \label{fig:nvs_compare}
       \vspace{-4mm}
\end{figure*}

\section{Experiments}

\subsection{Experimental Setting}

\noindent \textbf{Datasets.}
We conduct experiments on the Waymo Open Dataset~\cite{sun2020waymo}, which provides synchronized multi-view camera images and LiDAR scans captured in diverse real-world driving scenarios. 
Following the evaluation set in \cite{du2024dreamcar}, we randomly select $100$ vehicle instances that have been observed for more than $8$ views from the first $30$ scenes of the validation split.\footnote{The selected scene IDs and vehicle IDs will be public.} 
For each vehicle, we prepare $N=8$ images from different cameras or frames: $4$ are used as conditioning inputs, and the remaining $4$ are served as reversed views for image-based evaluation.  
The pseudo 3D ground-truth for geometry evaluation is derived by aggregating LiDAR scans in each frame and segmenting vehicle instances using the provided instance masks.

\noindent \textbf{Implementation Details.}
Our framework builds upon the pretrained TRELLIS~\cite{xiang2025trellis} without any additional finetuning. 
Stage~I predicts a sparse $64^3$ voxel occupancy grid.
Differently, during inference, we extend the sampling steps from $25$ to $100$, and adjust the classifier-free guidance (CFG) to $3.0$ Stage~I.
Moreover, voxel pose optimization is performed at the $30$th step, followed by $70$ steps of voxel-guided geometry generation using a BCE loss on LiDAR-occupied voxels.
For voxel filtering, we decode SLAT features into 3DGS and prune Gaussians with opacity threshold $\tau_\alpha=0.005$ before re-voxelization. 

\noindent \textbf{Baselines.}
We conduct comprehensive comparisons with state-of-the-art vehicle generation and 3D asset generation methods. Among them, DreamCar~\cite{du2024dreamcar} is the SOTA vehicle generation method since some more recent methods \cite{liu2025protocar} and \cite{lin2024drive123} are not open source; InstantMesh~\cite{xu2024instantmesh}, Trellis~\cite{xiang2025trellis} and Hunyuan3D-2.0~\cite{zhao2025hunyuan3d2.0} are several highly influential 3D generation methods. Moreover, we adopt MV-TRELLIS and MV-Hunyuan3D as the TRELLIS and Hunyuan3D model with multi-image conditioning. 
For fairness, InstantMesh and TRELLIS, which take a single image as input, are evaluated by averaging the results over four runs for the same $4$ conditioning images, as used in multi-view image methods.

\noindent \textbf{Metrics.}
We evaluate the generated 3D vehicles using both image- and geometry-level metrics. 
For image quality,  we adopt PSNR, SSIM~\cite{wang2004ssim}, and LPIPS~\cite{zhang2018lpips} to measure the fidelity of rendered novel views against ground-truth images.
For geometry, we compute Chamfer Distance (CD) between the generated mesh and aggregated LiDAR point clouds. Since LiDAR point clouds serve as pseudo ground truth and may be incomplete,
we report both the one-sided Chamfer Distance (CD$_{one}$) and bidirectional (CD$_{bi}$). CD$_{one}$ measures the distance from LiDAR points to the generated mesh and reflects how well the reconstruction matches the real-world shape. CD$_{bi}$ further includes mesh-to-LiDAR distance, capturing the issue of extra-generation, where higher scores indicate floaters or extra structures. 

\subsection{Comparison with State-of-the-art Methods}

\begin{figure*}[ht]
      \centering
      \includegraphics[width=0.95\linewidth]{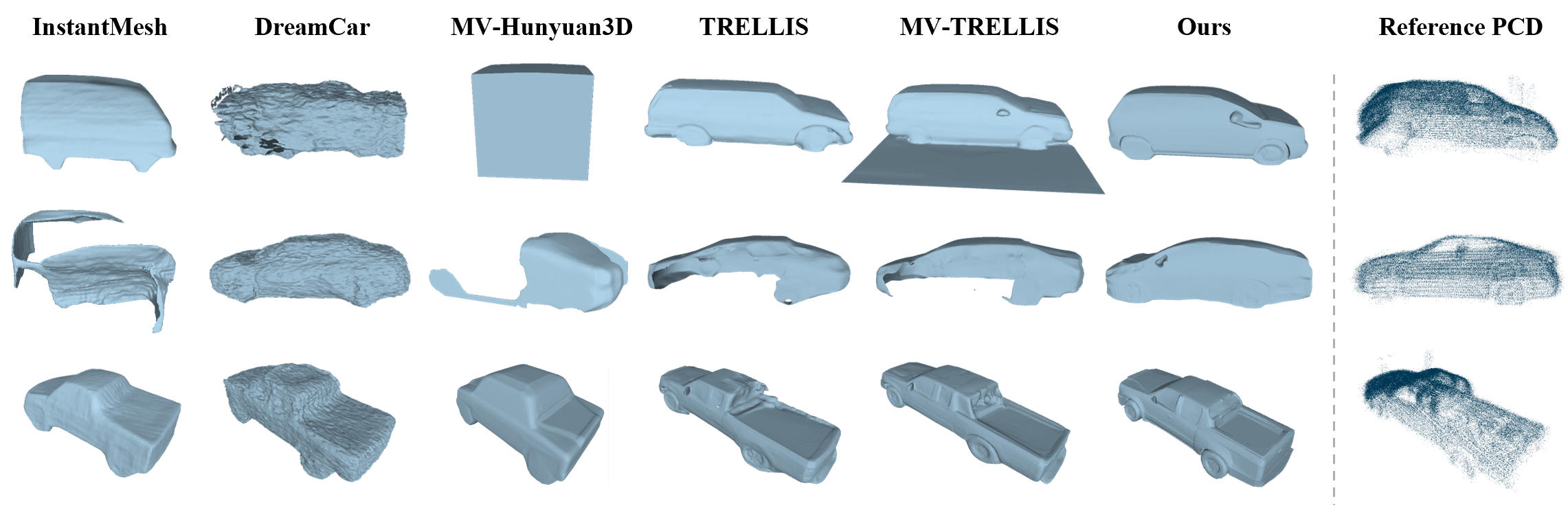}
      \caption{ 
      Qualitative comparison with baseline methods in 3D geometry.
      Rightmost column shows the reference LiDAR point cloud.
      }
      \label{fig:geo_compare}
\end{figure*}

\begin{figure*}[ht]
      \centering
      \includegraphics[width=0.85\linewidth]{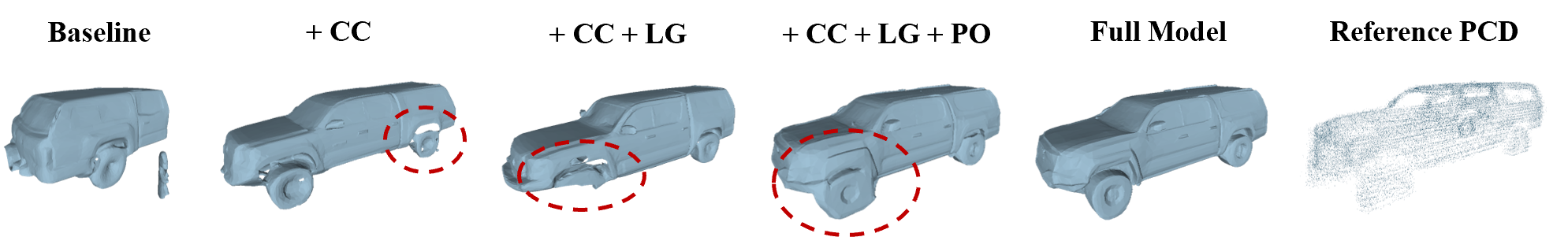}
      \caption{
      An visual comparison for ablation study.
       CC, LG, PO stand for cycle conditioning, LiDAR guidance, and pose optimization respectively. Red dashed circles highlight improvements.
      }
      \label{fig:ablation_vis}
       \vspace{-2mm}
\end{figure*}

\noindent \textbf{Evaluation on Novel-view Synthesis.}
We evaluate the results of novel view synthesis (NVS)  to mainly assess the geometric accuracy and generalization on real-world images of our method. The quantitative and qualitative results are presented in Table~\ref{tab:comp1} and Fig.~\ref{fig:nvs_compare}, respectively. As shown in Table 1, our method achieves best performance in both PSNR and LPIPS. Notably, despite reporting high PSNR and SSIM, InstantMesh~\cite{xu2024instantmesh} and MV-Hunyuan3D~\cite{hunyuan3d2025hunyuan3d2.1} often produce highly inaccurate geometry, as revealed by the Chamfer Distance metric and Fig.~\ref{fig:geo_compare}. Upon closer inspection, we find the failure cases of InstantMesh and MV-Hunyuan3D often produce severely inaccurate geometry and blurry textures. Since these meshes deviate significantly from true vehicle shape, they cannot be properly aligned in the canonical space. As a result, the rendering results only cover part of the vehicle with over-smoothed texture, which may be favored by PSNR and SSIM. In addition, since PSNR/SSIM are computed only on valid pixels, they fail to capture our method’s advantage in completing occluded regions.

Meanwhile, the comparison with TRELLIS~\cite{xiang2025trellis}, DreamCar~\cite{du2024dreamcar}, and MV-TRELLIS can more accurately reflect the geometry quality, as the orientation of their meshes align more faithfully with the true vehicle.
Specifically, when comparing with the TRELLIS series, it can be observed that the cars generated by TRELLIS tend to be slim, with inaccurate length-to-height ratios (see Fig.~\ref{fig:nvs_compare} first-second rows). In contrast, our results preserve correct proportions, enabling consistent pixel-alignment with ground-truth and yielding over 2dB PSNR improvement.

        \begin{table}[t]
            \centering
            \caption{ 
            Ablation study of the proposed modules.
            }
            \renewcommand{\arraystretch}{0.9}
             \label{tab:ablation}
             \setlength{\tabcolsep}{4pt}
                \begin{tabular}{c|cccc|cc}
                \toprule
                    ID &  \makecell{Cycle\\Cond.} & \makecell{LiDAR\\Gui.}  & \makecell{Pose\\Optim.} & \makecell{Mesh\\Refine} &  CD$_{one}\downarrow$ & CD$_{bi}\downarrow$   \\
                \midrule
                     1 &  &  &  & & 0.0848 & 0.1050 \\
                    2 &  \textcolor{black}{\ding{51}}  &  &  &   & 0.0388		& 0.0479 \\
                    3 &  \textcolor{black}{\ding{51}}  &  \textcolor{black}{\ding{51}}  &    &  & 0.0292 & 0.1526 \\
                    4 &   \textcolor{black}{\ding{51}}  &  \textcolor{black}{\ding{51}}  & \textcolor{black}{\ding{51}} & &  0.0267 &   0.0496  \\
                    5 (Ours) &  \textcolor{black}{\ding{51}} & \textcolor{black}{\ding{51}} &  \textcolor{black}{\ding{51}}  & \textcolor{black}{\ding{51}}    & \textbf{0.0210}		& \textbf{0.0361}  \\
                \bottomrule
            \end{tabular}
             \vspace{-4mm}
        \end{table}

\noindent \textbf{Evaluation on Mesh Generation.}
Similar conclusions emerge when evaluating Chamfer Distance (CD) to directly measure the geometric accuracy of the reconstructed meshes. As shown in Tab.~\ref{tab:comp1}, our method achieves the lowest error in both one-sided and bidirectional CD, significantly outperforming the TRELLIS series baseline. Combined with the visual comparisons in Fig.~\ref{fig:geo_compare}, we observe that single-view methods (InstantMesh~\cite{xu2024instantmesh} and TRELLIS~\cite{xiang2025trellis}) often produce incomplete or distorted results due to their strong reliance on the input image quality. DreamCar’s~\cite{du2024dreamcar} reconstructions are coarse, limited by its NeRF-based mesh extraction. MV-Hunyuan3D~\cite{hunyuan3d2025hunyuan3d2.1} struggles with arbitrary multi-view inputs, as it is trained under fixed viewpoints. MV-TRELLIS~\cite{xiang2025trellis}, while better leveraging multi-view conditioning, still suffers from unstable extra-mesh generation (Fig.~\ref{fig:geo_compare}-first row), incompleteness(Fig.~\ref{fig:geo_compare}-second row), and inaccurate metric length(Fig.~\ref{fig:geo_compare}-third row). In contrast, our proposed point cloud guidance effectively resolves these issues. Although it may seem unfair that we use LiDAR as condition, yet our ability to fully exploit this information is the core contribution of our method. Both quantitative and qualitative results demonstrate that we generate meshes that are highly consistent with real-world vehicles shape.

 In terms of efficiency, MM-TRELLIS reconstructs one vehicle in 37.3s on a single NVIDIA H20 GPU. The generation process (Stage I and Stage II) takes 23.6 s, followed by 13.7 s for mesh decoding.

\subsection{Ablation Study}
We conduct ablation experiments to validate the contribution of each component in our framework. As shown in Tab.~\ref{tab:ablation} and Fig.~\ref{fig:ablation_vis}, adding LiDAR guidance clearly improves the CD$_{one}$ score, indicating better alignment between LiDAR point cloud and the generated mesh, However, the orientation ambiguity can lead to unstable optimization, causing the CD$_{bi}$ becomes worse. Incorporating pose optimization effectively resolves this issue, leading to more stable training and more accurate geometry. Finally, the opacity-based mesh refinement further removes floaters and noisy artifacts, producing clean and realistic vehicle surfaces. 



{

\begin{figure}[ht]
      \centering
      \includegraphics[width=1.0\linewidth]{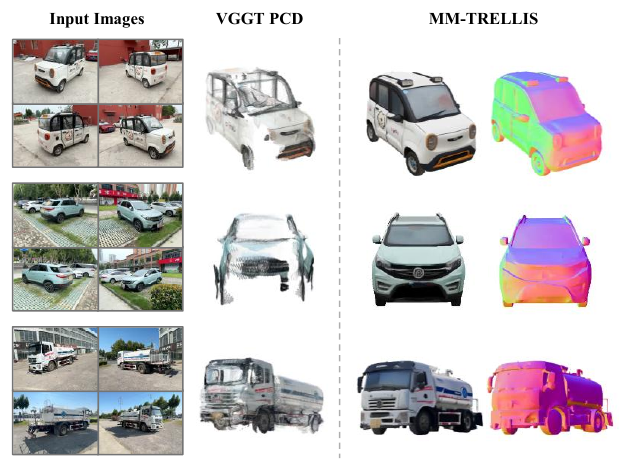}
      \caption{Extend our method to VGGT~\cite{wang2025vggt} point cloud as 3D priors. 
      Input images are sampled from the 3DRealCar~\cite{du20243drealcar} dataset, and point clouds reconstructed by VGGT are used as geometric guidance. 
      }
      \label{fig:VGGT_extension}
       \vspace{-2mm}
\end{figure}

\begin{figure}[ht]
      \centering
      \includegraphics[width=0.75\linewidth]{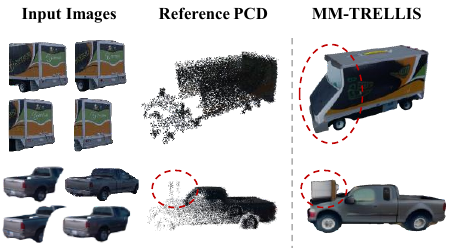}
      \caption{ Failure cases.
      }
      \label{fig:failiure_cases}
       \vspace{-2mm}
\end{figure}

\subsection{Extension to Feed-forward Point Cloud Priors}
Our method not only can utilize LiDAR point as priors, but also works on point cloud acquired from sparse-view images by other approaches (i.e. point clouds obtained from fast feed-forward reconstruction VGGT~\cite{wang2025vggt}).
As shown in Fig.~\ref{fig:VGGT_extension}, the reconstructed point clouds can be directly used as geometric guidance during voxel generation without any retraining. 
Although such point clouds are typically noisier and less complete than LiDAR data, MM-TRELLIS still produces geometrically consistent results. 
This demonstrates that our test-time voxel guidance mechanism is not restricted to LiDAR and can flexibly incorporate alternative 3D priors.

\subsection{Limitations}
As shown in Fig.~\ref{fig:failiure_cases}, our method may degrade when the inputs are severely incomplete or noisy. Truncated image observations (Fig.~\ref{fig:failiure_cases}-first column) can lead to missing geometry due to insufficient visual cues during cycle-conditioning. Noisy point clouds (Fig.~\ref{fig:failiure_cases}-second column)  may introduce incorrect geometric supervision, resulting in additional structures in the reconstructed mesh. 
Although opacity-based voxel filtering is applied, severe noise may still affect the final reconstruction.
These cases indicate that reliable visual cues and clean geometric guidance are important for stable reconstruction.

}

\section{CONCLUSIONS}

In this work, we present MM-TRELLIS, a zero-shot framework for real-world 3D vehicle generation that adapts native 3D diffusion priors to multimodal autonomous driving data. Our method combines multi-view cycle-conditioning with LiDAR-guided optimization to aggregate image information and enforce geometric consistency. A pose optimization is proposed to resolve orientation ambiguity, and an opacity-based filtering step is further applied to get a clean mesh. Experiments on the Waymo dataset show that MM-TRELLIS outperforms existing methods, delivering geometric-accurate and realistic 3D vehicle assets and highlighting its potential to enrich large-scale autonomous driving datasets.

\section*{ACKNOWLEDGMENT}

This work was supported by the National Natural Science Foundation of China (No. 62376282) and the Science and Technology Innovation Program of Hunan Province (No.2025RC3117).



\bibliographystyle{IEEEtran}
\bibliography{IEEEtranBST/ref}

@inproceedings{sun2020waymo,
  title={Scalability in perception for autonomous driving: Waymo open dataset},
  author={Sun, Pei and Kretzschmar, Henrik and Dotiwalla, Xerxes and Chouard, Aurelien and Patnaik, Vijaysai and Tsui, Paul and Guo, James and Zhou, Yin and Chai, Yuning and Caine, Benjamin and others},
  booktitle={Proceedings of the IEEE/CVF conference on computer vision and pattern recognition},
  pages={2446--2454},
  year={2020}
}

@inproceedings{xiang2025trellis,
  title={Structured 3d latents for scalable and versatile 3d generation},
  author={Xiang, Jianfeng and Lv, Zelong and Xu, Sicheng and Deng, Yu and Wang, Ruicheng and Zhang, Bowen and Chen, Dong and Tong, Xin and Yang, Jiaolong},
  booktitle={Proceedings of the Computer Vision and Pattern Recognition Conference},
  pages={21469--21480},
  year={2025}
}

@article{zhao2025hunyuan3d2.0,
  title={Hunyuan3d 2.0: Scaling diffusion models for high resolution textured 3d assets generation},
  author={Zhao, Zibo and Lai, Zeqiang and Lin, Qingxiang and Zhao, Yunfei and Liu, Haolin and Yang, Shuhui and Feng, Yifei and Yang, Mingxin and Zhang, Sheng and Yang, Xianghui and others},
  journal={arXiv preprint arXiv:2501.12202},
  year={2025}
}

@article{hunyuan3d2025hunyuan3d2.1,
  title={Hunyuan3D 2.1: From Images to High-Fidelity 3D Assets with Production-Ready PBR Material},
  author={Hunyuan3D, Team and Yang, Shuhui and Yang, Mingxin and Feng, Yifei and Huang, Xin and Zhang, Sheng and He, Zebin and Luo, Di and Liu, Haolin and Zhao, Yunfei and others},
  journal={arXiv preprint arXiv:2506.15442},
  year={2025}
}

@article{zhao2023michelangelo,
  title={Michelangelo: Conditional 3d shape generation based on shape-image-text aligned latent representation},
  author={Zhao, Zibo and Liu, Wen and Chen, Xin and Zeng, Xianfang and Wang, Rui and Cheng, Pei and Fu, Bin and Chen, Tao and Yu, Gang and Gao, Shenghua},
  journal={Advances in neural information processing systems},
  volume={36},
  pages={73969--73982},
  year={2023}
}

@article{zhang2024clay,
  title={Clay: A controllable large-scale generative model for creating high-quality 3d assets},
  author={Zhang, Longwen and Wang, Ziyu and Zhang, Qixuan and Qiu, Qiwei and Pang, Anqi and Jiang, Haoran and Yang, Wei and Xu, Lan and Yu, Jingyi},
  journal={ACM Transactions on Graphics (TOG)},
  volume={43},
  number={4},
  pages={1--20},
  year={2024},
  publisher={ACM New York, NY, USA}
}

@article{ye2025hi3dgen,
  title={Hi3dgen: High-fidelity 3d geometry generation from images via normal bridging},
  author={Ye, Chongjie and Wu, Yushuang and Lu, Ziteng and Chang, Jiahao and Guo, Xiaoyang and Zhou, Jiaqing and Zhao, Hao and Han, Xiaoguang},
  journal={arXiv preprint arXiv:2503.22236},
  volume={3},
  pages={2},
  year={2025}
}

@article{li2025sparc3d,
  title={Sparc3D: Sparse Representation and Construction for High-Resolution 3D Shapes Modeling},
  author={Li, Zhihao and Wang, Yufei and Zheng, Heliang and Luo, Yihao and Wen, Bihan},
  journal={arXiv preprint arXiv:2505.14521},
  year={2025}
}

@article{wu2025direct3ds2,
  title={Direct3D-S2: Gigascale 3D Generation Made Easy with Spatial Sparse Attention},
  author={Wu, Shuang and Lin, Youtian and Zhang, Feihu and Zeng, Yifei and Yang, Yikang and Bao, Yajie and Qian, Jiachen and Zhu, Siyu and Cao, Xun and Torr, Philip and others},
  journal={arXiv preprint arXiv:2505.17412},
  year={2025}
}

@article{lipman2022flowmatching,
  title={Flow matching for generative modeling},
  author={Lipman, Yaron and Chen, Ricky TQ and Ben-Hamu, Heli and Nickel, Maximilian and Le, Matt},
  journal={arXiv preprint arXiv:2210.02747},
  year={2022}
}

@article{lai2025hunyuan3d2.5,
  title={Hunyuan3D 2.5: Towards High-Fidelity 3D Assets Generation with Ultimate Details},
  author={Lai, Zeqiang and Zhao, Yunfei and Liu, Haolin and Zhao, Zibo and Lin, Qingxiang and Shi, Huiwen and Yang, Xianghui and Yang, Mingxin and Yang, Shuhui and Feng, Yifei and others},
  journal={arXiv preprint arXiv:2506.16504},
  year={2025}
}

@article{du2024dreamcar,
  title={DreamCar: Leveraging Car-Specific Prior for In-the-Wild 3D Car Reconstruction},
  author={Du, Xiaobiao and Sun, Haiyang and Lu, Ming and Zhu, Tianqing and Yu, Xin},
  journal={IEEE Robotics and Automation Letters},
  year={2024},
  publisher={IEEE}
}

@inproceedings{muller2022autorf,
  title={Autorf: Learning 3d object radiance fields from single view observations},
  author={M{\"u}ller, Norman and Simonelli, Andrea and Porzi, Lorenzo and Bulo, Samuel Rota and Nie{\ss}ner, Matthias and Kontschieder, Peter},
  booktitle={Proceedings of the IEEE/CVF conference on computer vision and pattern recognition},
  pages={3971--3980},
  year={2022}
}

@inproceedings{guo2024supnerf,
  title={Sup-nerf: A streamlined unification of pose estimation and nerf for monocular 3d object reconstruction},
  author={Guo, Yuliang and Kumar, Abhinav and Zhao, Cheng and Wang, Ruoyu and Huang, Xinyu and Ren, Liu},
  booktitle={European Conference on Computer Vision},
  pages={37--53},
  year={2024},
  organization={Springer}
}

@article{liu2024carstudio,
  title={Car-studio: learning car radiance fields from single-view and unlimited in-the-wild images},
  author={Liu, Tianyu and Zhao, Hao and Yu, Yang and Zhou, Guyue and Liu, Ming},
  journal={IEEE Robotics and Automation Letters},
  volume={9},
  number={3},
  year={2024},
  publisher={IEEE}
}

@inproceedings{shen2023gina3d,
  title={Gina-3d: Learning to generate implicit neural assets in the wild},
  author={Shen, Bokui and Yan, Xinchen and Qi, Charles R and Najibi, Mahyar and Deng, Boyang and Guibas, Leonidas and Zhou, Yin and Anguelov, Dragomir},
  booktitle={Proceedings of the IEEE/CVF conference on computer vision and pattern recognition},
  pages={4913--4926},
  year={2023}
}

@inproceedings{yang2025genassets,
  title={GenAssets: Generating in-the-wild 3D Assets in Latent Space},
  author={Yang, Ze and Wang, Jingkang and Zhang, Haowei and Manivasagam, Sivabalan and Chen, Yun and Urtasun, Raquel},
  booktitle={Proceedings of the Computer Vision and Pattern Recognition Conference},
  pages={22392--22403},
  year={2025}
}

@inproceedings{xu2023discoscene,
  title={Discoscene: Spatially disentangled generative radiance fields for controllable 3d-aware scene synthesis},
  author={Xu, Yinghao and Chai, Menglei and Shi, Zifan and Peng, Sida and Skorokhodov, Ivan and Siarohin, Aliaksandr and Yang, Ceyuan and Shen, Yujun and Lee, Hsin-Ying and Zhou, Bolei and others},
  booktitle={Proceedings of the IEEE/CVF conference on computer vision and pattern recognition},
  pages={4402--4412},
  year={2023}
}

@inproceedings{liu2025protocar,
  title={ProtoCar: Learning 3D Vehicle Prototypes from Single-View and Unconstrained Driving Scene Images},
  author={Liu, Hongyuan and Yu, Haochen and Zou, Bochao and Lyu, Juntao and Mei, Qi and Chen, Jiansheng and Ma, Huimin},
  booktitle={Proceedings of the AAAI Conference on Artificial Intelligence},
  volume={39},
  number={5},
  pages={5460--5468},
  year={2025}
}

@article{lin2024drive123,
  title={Drive-1-to-3: Enriching diffusion priors for novel view synthesis of real vehicles},
  author={Lin, Chuang and Zhuang, Bingbing and Sun, Shanlin and Jiang, Ziyu and Cai, Jianfei and Chandraker, Manmohan},
  journal={arXiv preprint arXiv:2412.14494},
  year={2024}
}

@article{viola2024marigold_dc,
  title={Marigold-dc: Zero-shot monocular depth completion with guided diffusion},
  author={Viola, Massimiliano and Qu, Kevin and Metzger, Nando and Ke, Bingxin and Becker, Alexander and Schindler, Konrad and Obukhov, Anton},
  journal={arXiv preprint arXiv:2412.13389},
  year={2024}
}

@inproceedings{deitke2023objaverse,
  title={Objaverse: A universe of annotated 3d objects},
  author={Deitke, Matt and Schwenk, Dustin and Salvador, Jordi and Weihs, Luca and Michel, Oscar and VanderBilt, Eli and Schmidt, Ludwig and Ehsani, Kiana and Kembhavi, Aniruddha and Farhadi, Ali},
  booktitle={Proceedings of the IEEE/CVF conference on computer vision and pattern recognition},
  pages={13142--13153},
  year={2023}
}

@article{deitke2023objaversexl,
  title={Objaverse-xl: A universe of 10m+ 3d objects},
  author={Deitke, Matt and Liu, Ruoshi and Wallingford, Matthew and Ngo, Huong and Michel, Oscar and Kusupati, Aditya and Fan, Alan and Laforte, Christian and Voleti, Vikram and Gadre, Samir Yitzhak and others},
  journal={Advances in Neural Information Processing Systems},
  volume={36},
  pages={35799--35813},
  year={2023}
}

@article{chang2015shapenet,
  title={Shapenet: An information-rich 3d model repository},
  author={Chang, Angel X and Funkhouser, Thomas and Guibas, Leonidas and Hanrahan, Pat and Huang, Qixing and Li, Zimo and Savarese, Silvio and Savva, Manolis and Song, Shuran and Su, Hao and others},
  journal={arXiv preprint arXiv:1512.03012},
  year={2015}
}

@article{kerbl20233dgs,
  title={3D Gaussian splatting for real-time radiance field rendering.},
  author={Kerbl, Bernhard and Kopanas, Georgios and Leimk{\"u}hler, Thomas and Drettakis, George},
  journal={ACM Trans. Graph.},
  volume={42},
  number={4},
  pages={139--1},
  year={2023}
}

@inproceedings{niemeyer2021giraffe,
  title={Giraffe: Representing scenes as compositional generative neural feature fields},
  author={Niemeyer, Michael and Geiger, Andreas},
  booktitle={Proceedings of the IEEE/CVF conference on computer vision and pattern recognition},
  pages={11453--11464},
  year={2021}
}

@inproceedings{chan2022eg3d,
  title={Efficient geometry-aware 3d generative adversarial networks},
  author={Chan, Eric R and Lin, Connor Z and Chan, Matthew A and Nagano, Koki and Pan, Boxiao and De Mello, Shalini and Gallo, Orazio and Guibas, Leonidas J and Tremblay, Jonathan and Khamis, Sameh and others},
  booktitle={Proceedings of the IEEE/CVF conference on computer vision and pattern recognition},
  pages={16123--16133},
  year={2022}
}

@article{karpikova2025madrive,
  title={MADrive: Memory-Augmented Driving Scene Modeling},
  author={Karpikova, Polina and Selikhanovych, Daniil and Struminsky, Kirill and Musaev, Ruslan and Golitsyna, Maria and Baranchuk, Dmitry},
  journal={arXiv preprint arXiv:2506.21520},
  year={2025}
}

@inproceedings{chen2025dora,
  title={Dora: Sampling and benchmarking for 3d shape variational auto-encoders},
  author={Chen, Rui and Zhang, Jianfeng and Liang, Yixun and Luo, Guan and Li, Weiyu and Liu, Jiarui and Li, Xiu and Long, Xiaoxiao and Feng, Jiashi and Tan, Ping},
  booktitle={Proceedings of the Computer Vision and Pattern Recognition Conference},
  pages={16251--16261},
  year={2025}
}

@article{zhang20233dshape2vecset,
  title={3dshape2vecset: A 3d shape representation for neural fields and generative diffusion models},
  author={Zhang, Biao and Tang, Jiapeng and Niessner, Matthias and Wonka, Peter},
  journal={ACM Transactions On Graphics (TOG)},
  volume={42},
  number={4},
  pages={1--16},
  year={2023},
  publisher={ACM New York, NY, USA}
}

@inproceedings{pooledreamfusion,
  title={DreamFusion: Text-to-3D using 2D Diffusion},
  author={Poole, Ben and Jain, Ajay and Barron, Jonathan T and Mildenhall, Ben},
  year = {2023},
  booktitle={The Eleventh International Conference on Learning Representations}
}

@inproceedings{lin2023magic3d,
  title={Magic3d: High-resolution text-to-3d content creation},
  author={Lin, Chen-Hsuan and Gao, Jun and Tang, Luming and Takikawa, Towaki and Zeng, Xiaohui and Huang, Xun and Kreis, Karsten and Fidler, Sanja and Liu, Ming-Yu and Lin, Tsung-Yi},
  booktitle={Proceedings of the IEEE/CVF Conference on Computer Vision and Pattern Recognition},
  pages={300--309},
  year={2023}
}

@inproceedings{qianmagic123,
  title={Magic123: One Image to High-Quality 3D Object Generation Using Both 2D and 3D Diffusion Priors},
  author={Qian, Guocheng and Mai, Jinjie and Hamdi, Abdullah and Ren, Jian and Siarohin, Aliaksandr and Li, Bing and Lee, Hsin-Ying and Skorokhodov, Ivan and Wonka, Peter and Tulyakov, Sergey and others},
  booktitle={The Twelfth International Conference on Learning Representations}
}

@inproceedings{chen2023fantasia3d,
  title={Fantasia3d: Disentangling geometry and appearance for high-quality text-to-3d content creation},
  author={Chen, Rui and Chen, Yongwei and Jiao, Ningxin and Jia, Kui},
  booktitle={Proceedings of the IEEE/CVF international conference on computer vision},
  pages={22246--22256},
  year={2023}
}

@article{wang2024prolificdreamer,
  title={Prolificdreamer: High-fidelity and diverse text-to-3d generation with variational score distillation},
  author={Wang, Zhengyi and Lu, Cheng and Wang, Yikai and Bao, Fan and Li, Chongxuan and Su, Hang and Zhu, Jun},
  journal={Advances in Neural Information Processing Systems},
  volume={36},
  year={2024}
}

@inproceedings{liu2023zero123,
  title={Zero-1-to-3: Zero-shot one image to 3d object},
  author={Liu, Ruoshi and Wu, Rundi and Van Hoorick, Basile and Tokmakov, Pavel and Zakharov, Sergey and Vondrick, Carl},
  booktitle={Proceedings of the IEEE/CVF international conference on computer vision},
  pages={9298--9309},
  year={2023}
}

@article{shi2023zero123++,
  title={Zero123++: a single image to consistent multi-view diffusion base model},
  author={Shi, Ruoxi and Chen, Hansheng and Zhang, Zhuoyang and Liu, Minghua and Xu, Chao and Wei, Xinyue and Chen, Linghao and Zeng, Chong and Su, Hao},
  journal={arXiv preprint arXiv:2310.15110},
  year={2023}
}

@inproceedings{long2024wonder3d,
  title={Wonder3d: Single image to 3d using cross-domain diffusion},
  author={Long, Xiaoxiao and Guo, Yuan-Chen and Lin, Cheng and Liu, Yuan and Dou, Zhiyang and Liu, Lingjie and Ma, Yuexin and Zhang, Song-Hai and Habermann, Marc and Theobalt, Christian and others},
  booktitle={Proceedings of the IEEE/CVF Conference on Computer Vision and Pattern Recognition},
  pages={9970--9980},
  year={2024}
}

@article{wu2024unique3d,
  title={Unique3D: High-Quality and Efficient 3D Mesh Generation from a Single Image},
  author={Wu, Kailu and Liu, Fangfu and Cai, Zhihan and Yan, Runjie and Wang, Hanyang and Hu, Yating and Duan, Yueqi and Ma, Kaisheng},
  journal={arXiv preprint arXiv:2405.20343},
  year={2024}
}

@article{xu2024instantmesh,
  title={Instantmesh: Efficient 3d mesh generation from a single image with sparse-view large reconstruction models},
  author={Xu, Jiale and Cheng, Weihao and Gao, Yiming and Wang, Xintao and Gao, Shenghua and Shan, Ying},
  journal={arXiv preprint arXiv:2404.07191},
  year={2024}
}

@article{liu2025stnerf,
  title={STNeRF: symmetric triplane neural radiance fields for novel view synthesis from single-view vehicle images},
  author={Liu, Zhao and Fu, Zhongliang and Li, Gang and Hu, Jie and Yang, Yang},
  journal={Applied Intelligence},
  volume={55},
  number={5},
  pages={322},
  year={2025},
  publisher={Springer}
}

@article{wang2004ssim,
  title={Image quality assessment: from error visibility to structural similarity},
  author={Wang, Zhou and Bovik, Alan C and Sheikh, Hamid R and Simoncelli, Eero P},
  journal={IEEE transactions on image processing},
  volume={13},
  number={4},
  pages={600--612},
  year={2004},
  publisher={IEEE}
}

@inproceedings{zhang2018lpips,
  title={The unreasonable effectiveness of deep features as a perceptual metric},
  author={Zhang, Richard and Isola, Phillip and Efros, Alexei A and Shechtman, Eli and Wang, Oliver},
  booktitle={Proceedings of the IEEE conference on computer vision and pattern recognition},
  pages={586--595},
  year={2018}
}

@inproceedings{ester1996dbscan,
  title={A density-based algorithm for discovering clusters in large spatial databases with noise},
  author={Ester, Martin and Kriegel, Hans-Peter and Sander, J{\"o}rg and Xu, Xiaowei and others},
  booktitle={kdd},
  volume={96},
  number={34},
  pages={226--231},
  year={1996}
}

@article{choi2025enhanced,
  title={Enhanced Diffusion Sampling via Extrapolation with Multiple ODE Solutions},
  author={Choi, Jinyoung and Kang, Junoh and Han, Bohyung},
  journal={arXiv preprint arXiv:2504.01855},
  year={2025}
}

@inproceedings{wang2025vggt,
  title={Vggt: Visual geometry grounded transformer},
  author={Wang, Jianyuan and Chen, Minghao and Karaev, Nikita and Vedaldi, Andrea and Rupprecht, Christian and Novotny, David},
  booktitle={Proceedings of the Computer Vision and Pattern Recognition Conference},
  pages={5294--5306},
  year={2025}
}

@article{du20243drealcar,
  title={3DRealCar: An In-the-wild RGB-D Car Dataset with 360-degree Views},
  author={Du, Xiaobiao and Sun, Haiyang and Wang, Shuyun and Wu, Zhuojie and Sheng, Hongwei and Ying, Jiaying and Lu, Ming and Zhu, Tianqing and Zhan, Kun and Yu, Xin},
  journal={arXiv preprint arXiv:2406.04875},
  year={2024}
}

@article{li2025refsam,
  title={Refsam: Efficiently adapting segmenting anything model for referring video object segmentation},
  author={Li, Yonglin and Zhang, Jing and Teng, Xiao and Zhang, Haoyu and Liu, Xinwang and Lan, Long},
  journal={Neural Networks},
  pages={108000},
  year={2025},
  publisher={Elsevier}
}

@article{xia2025d,
  title={{$D^2$GS}: Dense Depth Regularization for LiDAR-free Urban Scene Reconstruction},
  author={Xia, Kejing and Jia, Jidong and Jin, Ke and Bai, Yucai and Sun, Li and Tao, Dacheng and Zhang, Youjian},
  journal={arXiv preprint arXiv:2510.25173},
  year={2025}
}

@article{lan2026c,
  title={C-WOE: Clustering for Out-of-Distribution Detection Learning with Wild Outlier Exposure},
  author={Lan, Long and Hu, Zhaohui and Li, He and Liu, Tongliang and Liu, Xinwang},
  journal={IEEE Transactions on Image Processing},
  year={2026},
  publisher={IEEE}
}

\end{document}